\pdfoutput=1

\documentclass[11pt]{article}

\usepackage{naacl2021}

\usepackage{times}
\usepackage{latexsym}

\usepackage[T1]{fontenc}

\usepackage[utf8]{inputenc}

\usepackage{microtype}

\usepackage{times}
\usepackage{latexsym}

\usepackage{microtype}

\usepackage{graphicx}
\usepackage{booktabs} 
\usepackage{multicol}
\usepackage{makecell}
\usepackage{amsmath}
\usepackage{amssymb}
\usepackage[utf8]{inputenc}
\usepackage{fixltx2e}
\usepackage{hyperref}
\usepackage{amssymb}
\usepackage{booktabs}
\usepackage{tabularx}

\newcolumntype{P}[1]{>{\centering\arraybackslash}p{#1}}
\newcolumntype{M}[1]{>{\centering\arraybackslash}m{#1}}
\newcommand{\quotes}[1]{``#1''}
\usepackage{float}

\newcolumntype{L}{>{\centering\arraybackslash}m{4.5cm}}
\newcolumntype{R}{>{\centering\arraybackslash}m{6cm}}
\newcolumntype{Q}{>{\centering\arraybackslash}m{7.5cm}}

\title{LaTeX-Numeric: \underline{L}anguage-\underline{a}gnostic \underline{T}ext attribute \underline{eX}traction for E-commerce \underline{Numeric} Attributes}

\author{Kartik Mehta \\
  India Machine Learning \\
  Amazon \\
  \texttt{kartim@amazon.com} \\\And
  Ioana Oprea \\
  Retail Business Services \\
  Amazon \\
  \texttt{ioanao@amazon.com} \\\And
  Nikhil Rasiwasia \\
  India Machine Learning \\
  Amazon \\
  \texttt{rasiwasi@amazon.com} \\}

\begin{document}

\maketitle
\begin{abstract}
In this paper, we present LaTeX-Numeric - a high-precision fully-automated scalable framework for extracting E-commerce numeric attributes from 
product text like product description. Most of the past work on attribute extraction is not scalable as they rely on manually curated training
data, either with or without the use of active learning. We rely on distant supervision for training data generation, removing dependency on manual labels.
One issue with distant supervision is that it leads to incomplete training annotation due to missing attribute values while matching. We propose a 
multi-task learning architecture to deal with missing labels in the training data, leading to F1 improvement of 9.2\% for numeric attributes over 
single-task architecture. While multi-task architecture benefits both numeric and non-numeric attributes, we present automated
techniques to further improve the numeric attributes extraction models. Numeric attributes require a list of units (or aliases) for better matching
with distant supervision.  We propose an automated algorithm for alias creation using product text and attribute values, leading to a 20.2\% F1
improvement. Extensive experiments on real world dataset for 20 numeric attributes across 5 product categories and 3 English marketplaces show that LaTeX-numeric 
achieves a high F1-score, without any manual intervention, making it suitable for practical applications. Finally, we show that the improvements are 
language-agnostic and LaTeX-Numeric achieves 13.9\% F1 improvement for 3 Romance languages\footnote{https://www.britannica.com/topic/Romance-languages}.
\end{abstract}

\section{Introduction}
\label{submission}
E-commerce websites often sell billions of products. These websites provide information in form of product images, product text 
(such as title and product description) and structured information, henceforth, termed as product attributes\footnote{E.g. RAM, weight and front\_camera are
some of the product attributes for mobile phone. We use the terminologies `product attributes' and `attributes' interchangeably in this paper.}. These attributes often act as a concise
summary of product information and are useful in product discovery, comparison and purchase decisions. They are usually
provided by selling partners at the time of product listing and can be missing or invalid, even though they might be present in product text sources. 
Extracting attribute values from these product text sources can be used to populate the missing attribute values and is the focus of this work.

Attribute Extraction from free form text can be posed as a Named Entity Recognition (NER) problem~\cite{zheng2018opentag}. Recently, deep learning 
models ~\cite{lample2016neural, ma2016end, huang2015bidirectional} have shown remarkable performance on NER tasks, eliminating the need of manually 
curated features. However, these approaches still require large amount of labelled data. While active learning can be used to efficiently curate
training data~\cite{zheng2018opentag}, however gathering data for hundreds of product categories and attributes is a resource extensive
task. One solution is to use distant supervision to create training data. Distant supervision has been extensively used to curate training 
set without manual effort for relation extraction~\cite{mintz2009distant}. In context of attribute extraction for
E-commerce, we can curate training data by using attribute values and matching them with tokens in product text. However, if attribute 
values are missing, distant supervision leads to missing annotations, a phenomenon not studied in literature.

In this work, we present an automated framework for building high-precision attribute extraction models for numeric attributes using distant supervision.
Multiple works in literature~\cite{madaan2016numerical, ibrahim2016making}  argue that distant supervision for numeric attributes poses unique challenges
and have given separate treatment to numeric attributes. Highlighted below are some interesting challenges that distant supervision poses for numeric 
attribute extraction models:\\
\textbf{Partial Annotations:} Distant supervision leads to incorrect annotations when attribute is present in the text field but structured attribute value is missing.\\
\textbf{Diverse surface forms:} There are multiple ways that attributes are mentioned in product text (e.g. resolution of `2' can be mentioned as `2 mp', `2 mpix' or `2 megapixels'). We term these different surface forms as alias.\\
\textbf{Confusing attributes:} Many attributes have common units and may have confusing mention in the text (e.g. `16 GB memory' refers to RAM while `128 GB memory storage' refers to `Hard Disk')\\
\textbf{Use of different units:} Seller may use diverse units for numeric attributes (e.g. `1.5 kg' as attribute value and `3.3 pounds' in product text).\\
Addressing these challenges in automated manner is the primary focus of this work. Our paper has the following contributions: (1) We propose a multi-task architecture to deal with partial annotations introduced due to missing attributes.
This multi-task architecture leads to F1 improvement of 9.2\% for numeric and 7.4\% for non-numeric attributes over single task 
architecture. (2) We propose a fully automated algorithm for alias creation using product text and attribute values. These alias improve 
the quality of training annotation in distant supervision, leading to models with 20.2\% F1 improvement for numeric attributes.
We demonstrate the effectiveness of our proposed approach using a real-world dataset of 20 numeric attributes across 5 categories and 3 English marketplaces.
Models trained using our proposed framework achieve a high F1-score without any manual intervention, 
making them suitable for practical applications. We show that our proposed approach is language agnostic. Experiments of using our 
framework on 3 Romance languages show 13.9\% F1 improvement. 
To the best of our knowledge, this is first successful attempt at building automated attribute extraction for numeric attributes at E-commerce 
scale. Rest of the paper is organized as follows. We describe our proposed framework and its components in Section~\ref{latex}. We describe the `Multi Task' architecture in Section~\ref{MultiAttributeExtraction} and `automated alias
creation' component in Section~\ref{aliascreation}. We describe datasets, experimental setup and results in 
Section~\ref{Evaluation}. Lastly, we summarize our work in Section~\ref{conclusion}.

\section{Related Work}

\subsection{Attribute Extraction for E-commerce}

Early works on information extraction focused on extracting facts from generic web pages 
~\cite{etzioni2005unsupervised, yates2007textrunner, etzioni2008open}. With rise of E-commerce, multiple works focused on extracting
attributes from product pages. ~\citet{ghani2006text} proposed use of supervised learning to extract attributes from E-commerce product
descriptions. ~\citet{putthividhya2011bootstrapped} formulated attribute extraction from short titles as NER problem, using
multiple base classifiers and a CRF layer. The training data was created by matching entries from a seed 
dictionary. ~\citet{more2016attribute} proposed use of distant supervision for attribute extraction. 
They used token-wise string matching (henceforth referred as exact match) based on attribute values to annotate title tokens and train an
NER model with manually defined features. They used manual intervention to improve the training annotations e.g. dealing with spelling 
mistakes and different surface forms of brand. ~\citet{majumder2018deep} extended 
this work with use of recurrent neural networks, excluding use of manually defined features. ~\citet{zheng2018opentag} proposed OpenTag 
using bidirectional LSTM, Conditional Random Fields (CRF) and attention mechanism. But the training data for OpenTag is manually created with use of active learning, 
making it challenging to use at E-commerce scale.

\subsection{Distant supervision of numeric attributes}
Getting manual training data has always been a resource intensive and expensive task and distant supervision has been explored as an alternative. 
Distant supervision for numeric attributes has been used for relation extraction ~\cite{hoffmann2010learning, madaan2016numerical}, 
question answering~\cite{davidov2010extraction}, entity linking~\cite{ibrahim2016making}.
~\citet{madaan2016numerical} argued that distant supervision for numerical attributes presents peculiar challenges not found for non-numeric
attributes, such as high noise due to matching out of context, low recall due to different rounding level, and importance of units.
~\citet{ibrahim2016making} constructed a KB from freebase.com, keeping a list of units and conversion rules for numeric quantities. While these works have 
established the importance of units for distant supervision of numeric attributes, but the list of units is manually curated.

\subsection{NER with partial annotation}
Distant supervision may lead to noisy training data due to partial annotations. ~\citet{tsuboi2008training} argued that partial annotations
may happen due to ambiguous annotation and proposed CRF-PA to alleviate the issue of partial annotations. ~\citet{yang2018distantly} studied 
partial annotations introduced due to incomplete dictionary and extended CRF-PA to NER models. \citet{jie-etal-2019-better} proposed learning the probability distribution of 
all possible label sequences compatible with given incomplete annotation, and using this probability to clean the training annotations. 
For E-commerce attribute extraction, partial annotations may happen due to missing attribute value. Our paper is the first work to establish this
phenomenon for attribute extraction and provide a systematic way to alleviate this problem. We compare our proposed approach with \citet{jie-etal-2019-better} 
in Section~\ref{mamt-eval}.

\section{LaTeX-Numeric Framework}
\label{latex}

\begin{figure}
	\includegraphics[width=\columnwidth]{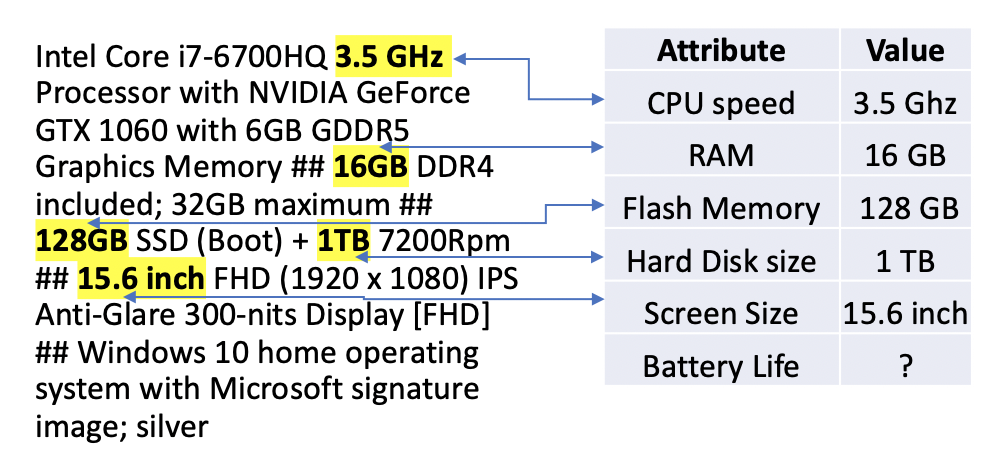}
	\caption{Illustration of E-commerce attribute extraction problem.}
	\label{fig:bp_extract}
\end{figure}

We pose the attribute extraction problem (refer Figure~\ref{fig:bp_extract} ) as a NER problem, where product attributes are treated as named entities. 
Formally, we are given a text $X$ with a particular tokenization ($x_1$, $x_2$,.....$x_m$) and a set of attributes $A$: ($\alpha_1$, 
$\alpha_2$..... $\alpha_n$). The task is to extract $v_i$ $=$ $\alpha_k$ for i $\in$ [1, m] where k $\in$ [0, n] and $\alpha_0$ represents `Other'
entity.

\begin{figure}
	\includegraphics[width=\columnwidth]{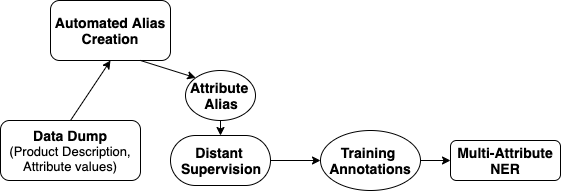}
	\caption{LaTeX-Numeric framework for extraction of E-commerce numeric attributes.}
	\label{fig:numtax}
\end{figure}

Figure~\ref{fig:numtax} gives an overview of our proposed \textit{LaTeX-Numeric} framework. We are given a list of pre-defined numeric attributes,
dump of products consisting of product text and existing attribute values. The attribute values are decimals (e.g. 16)
and have an underlying unit (e.g. GB). We term this underlying unit as the canonical unit. For creating distant supervision-based training annotations, 
we use these canonical units and combine them with attribute values for matching with product text. This serves as the `canonical aliasing' baseline for our comparisons. 
We use BIO tagging scheme for our experiments as it is a popular format. For training, we use the recently proposed BiLSTM-CNN-CRF model~\cite{ma2016end}. This model consists of CNN architecture to 
encode character information, LSTM-based encoder to model contextual information of each token and a CRF based tag decoder, which exploits 
the labels of neighboring tokens for improved classification. Unlike OpenTag~\cite{zheng2018opentag}, we don't use attention as we didn't observe any 
improvements with use of attention in our initial experiments.

Creating training annotations using distant supervision may lead to partial annotations due to missing attribute values. 
In Section~\ref{MultiAttributeExtraction}, we describe our proposed multi-task learning architecture to deal with such partial 
annotations. Additionally, we have observed that sellers use multiple surface forms to mention attributes in product text 
(e.g. `3mp', `3mpix', `3 megapixels' for resolution) and hence, distant supervision with just canonical units (e.g. `mp') may lead to 
suboptimal training annotations.  Curating a list of these diverse surface forms will help improve the quality of training annotations. 
We describe an automated approach for curating such diverse units and improving training annotations of numeric attributes in Section~\ref{aliascreation}.

\subsection{Multi Attribute Joint Extraction}

\label{MultiAttributeExtraction}
\begin{figure*}[t]
	\includegraphics[width=1.9\columnwidth]{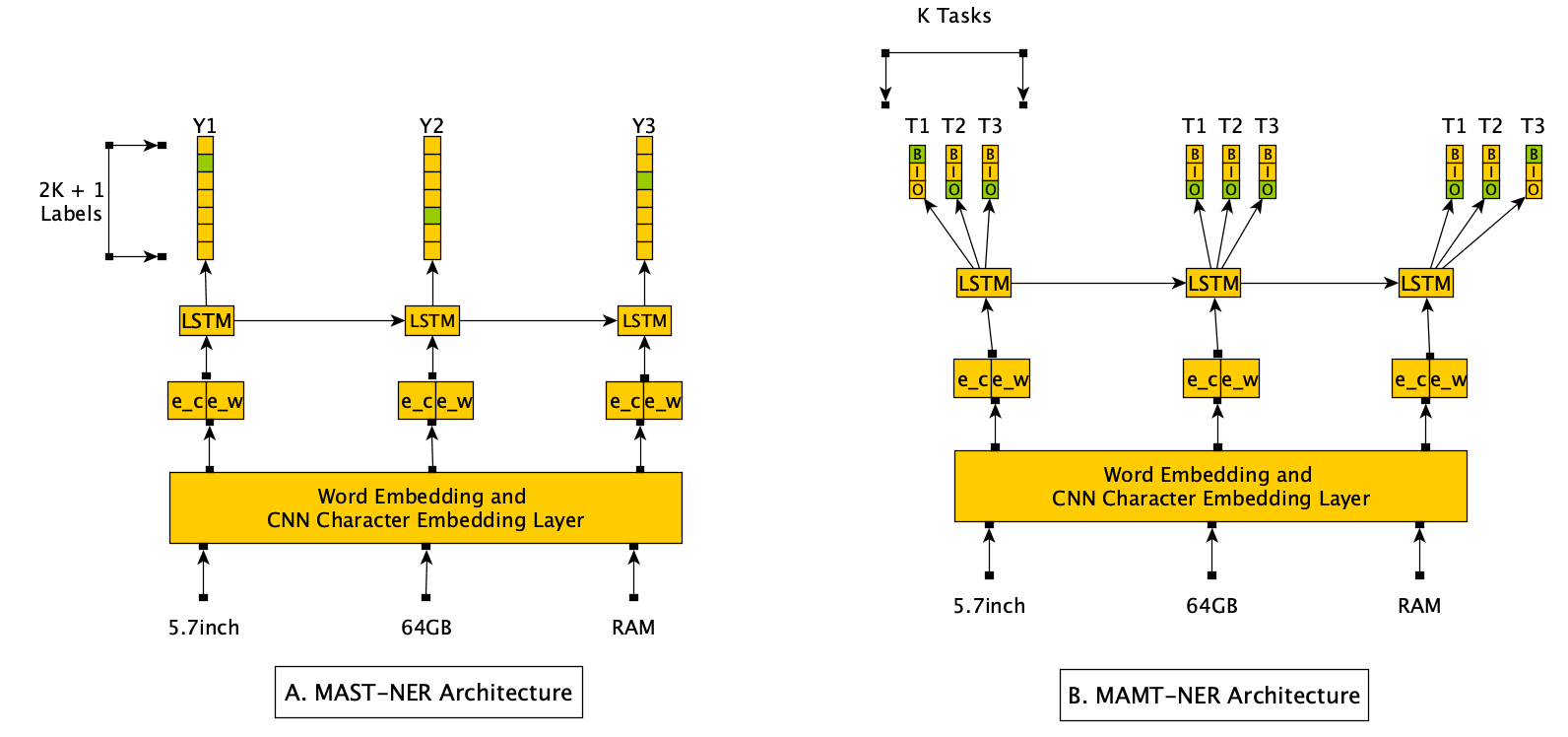}
	\caption{Figure showing different architectures for Multi Attributes Extraction models. We assume BIO-tagging of attributes with only 3 tags possible - B, I and O. For MAST-NER, we have two possible tags for each attribute and one others tag. For MAMT-NER, each attribute extraction is considered a separate task with weights shared for character embeddings, word embeddings and BiLSTM layer}
	\label{fig:NERArch}
\end{figure*}

To jointly extract multiple attributes, the tagging strategy can be modified to consider an output label with tags for all attributes. With `BIO'
tagging, each attribute has its own (`B' and `I') tags with `O' common for all attributes, leading to total $2K+1$ tags for $K$ attributes. Based
on this modified tagging, a single NER model can be trained for multi-attributes extraction. We term this setting of training `Multi Attribute
Single Task' model as MAST-NER (refer Figure~\ref{fig:NERArch}). MAST-NER is the commonly used strategy for attributes extraction~\cite{zheng2018opentag, sawant2017fashion, shen2017deep, joshi2015distributed}.

Under distant supervision, attribute value is used to find matches in product text tokens. However, if the attribute value is missing, 
no match will be found even when attribute value is mentioned in text and hence, the corresponding tokens are incorrectly tagged as `O'. 
We term this partial annotation due to missing attribute as Missing-PA problem. Table~\ref{tab:missingPA} shows an illustration of this problem. Missing-PA is generic to distant supervision
for multi-attributes and exists for non-numeric attributes as well. To the best of our knowledge, this problem has not received attention
in literature.

\begin{table}[H]
	\centering
	\small
	\begin{tabular}{r r r r r}
		\toprule
		 & \textbf{Display} & \textbf{RAM} & \textbf{Weight} & \textbf{BatteryLife}\\
		\toprule
		\makecell{Attribute \\ Value} & 12.3 & 16 & missing & 10\\
		\toprule
		\makecell{Canonical \\ Unit} & inches & gb & kg & hours\\
	\end{tabular}
\end{table}
\vspace{-2.2em}
\begin{table}[H]
	\centering
	\small
	\begin{tabular}{Q}
		\toprule
		The high performance Chromebook. Features 7th Gen Intel Core i7 processor, \textbf{\underline{16 GB}} RAM and 512 GB for storage. 
		The long lasting battery provides up to \textbf{\underline{10 hours}} of use and its fast charging so you can get 2 hours of use in 15 
		minutes \#\# Pixelbook's super thin and lightweight design measures 10.3 mm and weighs \textbf{\underline{1.2 kg}} Features a 
		\textbf{\underline{12.3 inches}} 360 touchscreen display\\
		\toprule
	\end{tabular}
	\caption{Illustration of Missing-PA for distant supervision. 1.2kg will be incorrectly tagged as `O' as value for weight attribute is missing.}
	\label{tab:missingPA}
\end{table}

To alleviate Missing-PA problem, one can train separate models for each attribute, by excluding samples where the corresponding 
attribute value is missing. However, such an approach requires training and managing a large number of models and separate 
computation for each attribute at evaluation time. Due to these practical challenges, this strategy is not suitable for practical
applications. Another way to alleviate missing-PA is to use MAST-NER setting, and to exclude all samples where atleast one attribute has
missing value. However, this approach may significantly reduce the size of training data as some attributes may have high missing rate,
leading to a suboptimal model. To alleviate this problem, we propose a multi-task learning architecture with separate output layers for each attribute as separate
tasks. We term this architecture of training `Multi Attribute Multi Task' model as MAMT-NER (Refer Figure~\ref{fig:NERArch}). MAMT-NER 
consists of shared character encoder, word encoder and BiLSTM layers. For each training sample, loss is deactivated (using masking) for tasks where 
corresponding attribute value is missing and activated only for remaining tasks where corresponding attribute values are non-missing. Loss for all activated 
tasks are weighted uniformly and weights of those tasks (including shared weights) are updated for the given sample. Note that the proposed MAMT-NER architecture is 
generic and can be used for non-numeric attributes as well as any underlying NER architecture, including recently proposed BERT~\cite{devlin2018bert}.

\subsection{Automated Alias Creation}
\label{aliascreation}
\vspace{-1mm}

\begin{figure}
	\centering
	\includegraphics[width=0.4\textwidth]{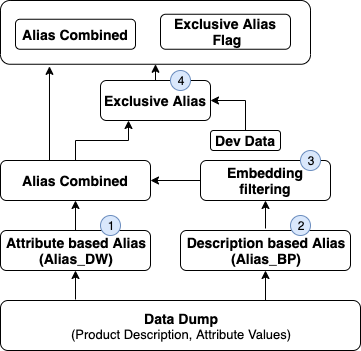}
	\caption[short caption]{Flow-diagram for Automated alias creation.}
	\label{fig:aliascreate}
\end{figure}

As argued earlier, the canonical unit is often not sufficient to capture diverse surface forms that sellers use to mention attributes in product text.
E.g. `13 inch', `13 inches', `13 in', are multiple ways to mention display\_size. One can analyze the mention of attribute values in product text and 
leverage that to create a list of commonly used surface forms. While, such an algorithm will detect common surface forms, it will miss out on units which
require a multiplicative factor (e.g. `pounds', `lbs' and `ounces' for weight where attribute values are in `kg'). To detect such units, we can analyze
all numeric mentions in product texts (in isolation from attribute value) and filter out noisy candidates by using similarity with canonical 
units in embedding space. Additionally, we have observed that some numeric attributes have units which are specific to those attributes 
(e.g. `mah' for battery\_power and `hertz' for refresh\_rate). One can detect such attributes and use this information while creating training annotations 
using distant supervision. Based on these learnings, we propose an approach for generating a more exhaustive list of aliases, in an automated fashion (Figure~\ref{fig:aliascreate}).

\subsubsection{Creation of alias\_dw}
\label{sec:aliasDW}

We create attribute-specific alias\_dw using product text and attribute values. 
We use a regex matching function, $\mathcal{M}$\footnote{$\mathcal{M} = re.findall(``\;" \;+\; $<$value$>$ \;+\; r\quotes{[\;-]*[a-zA-Z]+\;}, \;$<$text$>$)$, 
where $<$value$>$ is attribute value (e.g. 8 for RAM) and $<$text$>$ is product text.}, to
find candidate alias which are preceded by numeric attribute value in text. Though this matching function may lead to instances of collision (e.g. \textit{5} for RAM attribute may match with \textit{5 ghz} in text), we ignore cases where more than one match is found in text, to prevent impact of collisions. Alias\_dw leads to common surface form of attribute units (e.g. `in', `inches', `inch' for display\_size and `gb', `gigabyte' for hard\_disk).

\subsubsection{Creation of alias\_bp}
\label{sec:aliasBP}

We create a single alias\_bp (common across attributes) using product text data. We use a regex function $\mathcal{F}$\footnote{$\mathcal{F} = 
re.findall(``\;" \;+\; r\quotes{[\backslash d\backslash .]*[\backslash 
d][\;-]*[a-zA-Z]+\;}, \;$<$text$>$)$, where $<$text$>$ is product text.} to find candidate alias, which 
are tokens preceeded by any numeric mention in product text. Alias\_bp may contain noisy candidates which are not units for any attribute. We use word embeddings to match alias\_bp candidates with attributes and exclude noisy candidates.

\subsubsection{Embedding based filtering}
\label{embeddingfilter}
To remove noisy candidates and match alias\_bp candidates to attributes, we leverage canonical units and Glove embeddings. For each attribute, 
we calculate similarity of each alias\_bp candidate with its canonical unit in embedding space and keep only those candidates where similarity 
is greater than a pre-determined threshold. Thus, we obtain alias\_bp\_filter, which is attribute specific. E.g. we filter out `inches' and select 
`pounds' and `lbs' for weight attribute having `kg' as canonical unit.

Alias\_dw and alias\_bp\_filter complement each other. Alias\_bp\_filter misses out on units which have low similarity using embeddings (e.g. `in' for display\_size as `in' has a low similarity with `inch'). 
Alternately, alias\_dw misses out on cases where the unit mentioned in product text may require a multiplicative factor (e.g. alias\_dw for item\_weight misses out on `pounds' and `lbs'). 
We concatenate alias\_dw and alias\_bp\_filter to obtain alias\_combined for each attribute (shown for four attributes in Table~\ref{tab:samplealias}). 

With small manual effort, one can get the multiplicative factor for converting values in canonical units to units in alias\_combined and vice versa, which can further improve training annotations. 
As focus of current work is to build a fully automated attribute extraction system, we leave this as future work to be explored.

\begin{table}[H]
	\centering
	\small
	\begin{tabular}{c c r}
		\toprule
		Category & Attribute & Alias\_combined \\
		\toprule
		Laptop & Hard\_Disk & \makecell{[gb, mb, gigabyte, tb, x]}\\
		\toprule
		Laptop & Display & \makecell{[inches, inch, mm, cm, \\ft, centimeter, feet, in]}\\
		\toprule
		Tablet & Weight & \makecell{[kilograms, kgs, kg, grams, \\lbs, pounds, ounces, g]}\\
		\toprule
		TV & Refresh\_Rate & [hertz, hz]\\
		\toprule
	\end{tabular}
	\caption{Alias values shown for few attributes}
	\label{tab:samplealias}
\end{table}

\subsubsection{Exclusive Alias Flag}
\label{exclusivealias}

We use a small manually labelled dev set (created for hyper-parameters tuning) to create a flag indicating which attributes have exclusive alias.
We evaluate precision of extracting any mention of alias for a given attribute and if this precision is above a threshold, we consider that attribute 
to have exclusive alias. For attributes having an exclusive alias, we use regex-based matching for training annotations, tagging any numeric value followed
by the corresponding unit, irrespective of the attribute value. 

We refer our proposed approach of using `alias\_combined' and `exclusive alias flag' for creating training data of numeric attributes as `auto-aliasing' henceforth. 
We discuss experiments of using `auto-aliasing' as compared to other distant supervision techniques for numeric attributes in Section~\ref{matching-eval}.

\section{Experimental Setup and Results}
\label{Evaluation}
We picked five product categories and their 20 numeric attributes for three English marketplaces (IN, US and UK).
We extracted product data (product description and attribute values) for these categories and split this data into two parts (80\% train and 20\% test). 
The train part is used for automated alias creation and creation of training annotations with distant supervision. From the test part, we randomly 
picked products for each category and label the mention of category-specific numeric attributes in text. Out of the total labelled attribute-product pairs, we observed mention of 6.9K attributes in product text. We term training data for
English as `Train-EN' and audited test dataset as `Test-EN' (details in Table~\ref{table:GDS-EN}). To evaluate applicability of our proposed LaTeX-Numeric
framework for non-English languages, we did a similar analysis with one product category for three Romance languages of French (FR), Spanish (SP) and Italian (IT). 
We term this training data as `Train-Romance' and audited test dataset as `Test-Romance'. 
Similar to~\cite{zheng2018opentag}, we use F1-score for evaluation. Predictions are given full credit if correct value is extracted, but extracting 
more values than actual is considered incorrect (e.g. for a mobile phone with `4 gb' RAM, extracting either `4' or `4 gb' is considered correct prediction, 
but extracting two values of `4 gb' and `16 gb' is considered incorrect prediction).

\begin{table}
	\centering
	\small
	\begin{tabular}{M{15mm} r r r r}
		\toprule
		\makecell{Product\\ Category} & IN & US	& UK \\
		\toprule
		A (6)	& \makecell{14K (854)} & \makecell{241K (966)} & \makecell{70K (468)} \\
		B (4)	& \makecell{11K (396)} & \makecell{57K (493)} & \makecell{171K (320)} \\
		C (6) &  \makecell{43K (1027)} & \makecell{112K (623)} & \makecell{76K (559)} \\
		D (2) &  \makecell{3K (201)} & \makecell{14K (146)} & \makecell{32K (90)} \\
		E (2) &  \makecell{11K (246)} & \makecell{89K (266)} & \makecell{41K (244)} \\
		Total (20) &  \makecell{83K (2724)} & \makecell{514K (2494)} & \makecell{391K (1681)} \\
		\toprule
		\makecell{Product\\ Category} & FR & IT	& ES \\
		\toprule
		A (6)	& \makecell{93K (727)} & \makecell{70K (508)} & \makecell{74K (559)} \\
		\toprule
	\end{tabular}
	\caption{Stats for training and test data. Number of training products are shown with unit `K' (K=1000) and number of labelled attributes mention in test data is shown in adjacent parenthesis. Number of attributes is shown in parenthesis adjacent to each category.}
	\label{table:GDS-EN}
\end{table}

\subsection{Evaluation of Matching Techniques}

\label{matching-eval}

\begin{table}
	\centering
	\small
	\begin{tabular}{M{2.6cm} c c c c}
		\toprule
		\makecell{Matching \\Technique}	& IN & US & UK & Avg\\
		\toprule
		exact match												& 78.0				& 86.5		& 93.3		& 85.5\\
		\toprule
		canonical aliasing										& 100.0				& 100.0		& 100.0		& 100.0\\
		\toprule
		auto aliasing (our)	& \textbf{113.1} & \textbf{120.3} & \textbf{128.5} & \textbf{120.2}\\
		\toprule
	\end{tabular}
	\caption{Comparison of various matching techniques for training data generation using distant supervision (all numbers are relative to using canonical units).}
	\label{resultsmatching:tab}
\end{table}

In this section, we study improvements with our proposed alias creation. For comparison, we use two 
baselines of creating training annotation a) using lexical matching of numeric attribute value and product text (`exact match'), and b) matching based on canonical units (`canonical aliasing'). For each strategy, we use CNN-BiLSTM-CRF with 
MAST-NER architecture. Table~\ref{resultsmatching:tab} shows F1 score for using different matching techniques. `Canonical aliasing' approach 
shows better F1 score than `exact match', but it still suffers from low recall due to missing out on different surface 
forms mentioned in product text. With our proposed auto-aliasing, we address this limitation
of `canonical aliasing' and observe an average F1 improvement of 20.2\%, establishing `auto-aliasing' as best technique 
for distant supervision of numeric attributes. We use the training data created using `auto-aliasing' for all subsequent experiments (unless otherwise specified).

\subsection{Evaluation of MAMT Architecture }

\label{mamt-eval}
In this section, we perform a quantitative evaluation of our proposed MAMT architectures. Table~\ref{NERArchitecture:tab} shows results of using MAST and MAMT architecture 
with CNN-BiLSTM-CRF. As compared to MAST-NER architecture, we observe 9.2\% F1 improvement with our proposed MAMT architecture. Additionally, we observe that
\citet{jie-etal-2019-better}~\footnote{We use implementation of https://github.com/allanj/ner\_incomplete\_annotation. We show results 
only for IN as we get memory error when training for US and UK datasets which have larger training size.} shows better F1 score than MAST
due to higher recall. However, \citet{jie-etal-2019-better} leads to drop in precision due to confusion between close attributes (e.g. front-camera 
and back-camera for mobile.) Our proposed MAMT shows 8.7\% better F1 score than \citet{jie-etal-2019-better} for IN.

Further, we do comparison of MAST and MAMT architectures with BERT\footnote{We use implementation of https://github.com/namisan/mt-dnn, which uses bert-base and softmax as output layer.}
as underlying model and observed 3.5\% F1 improvement with MAMT, demonstrating its applicability to multiple underlying models. To establish the
effectiveness of MAMT architecture for non-numeric attributes, we curated a test dataset of 600 samples per attribute for 8 textual attributes across 4 product categories. As shown in Table~\ref{textualNERArchitecture:tab}, 
we observe 7.4\% F1 improvement on this dataset with our proposed MAMT-NER architecture, showing its effectiveness on textual attributes as well.

\begin{table}
\centering
\small
\begin{tabular}{M{30mm} c c c c}
\toprule
\makecell{Model + \\Architecture} & IN & US & UK & Avg\\
\toprule
BiLSTM-MAST	                    & 113.1             & 120.3             & 128.5       & 120.2\\
BiLSTM (\citet{jie-etal-2019-better})	                    & 114.4             & NA                & NA          & NA\\
BiLSTM-MAMT                       & 124.4             &  131.4            & 139.0       & 131.2\\
\toprule
BERT-MAST	                            & 117.7             & 122.9             & 128.8      & 122.8\\
BERT-MAMT                             &  120.4            & 127.8             & 134.3      & 127.1\\
\toprule
\end{tabular}
\caption{Study of multi-task architecture for numeric attributes. BERT uses softmax as output layer, while, BiLSTM refers to CNN-BiLSTM model with crf as output layer. All numbers are relative to using canonical units in Table~\ref{resultsmatching:tab}.}
\label{NERArchitecture:tab}
\end{table}

\begin{table}
\centering
\small
\begin{tabular}{M{2cm} c c c c}
\toprule
Model & \makecell{Archi-\\tecture} & Precision & Recall & F1 \\
\toprule
BiLSTM & MAST	     & 100.0 & 100.0 & 100.0\\
BiLSTM & MAMT        & 93.6 & 117.8 & \textbf{107.4}\\
\toprule
\end{tabular}
\caption{Study of multi-task architectures for textual attributes (all numbers are relative). BiLSTM refers to CNN-BiLSTM model with crf as output layer.}
\label{textualNERArchitecture:tab}
\end{table}

\subsection{Evaluation on non-English Languages}

\label{nonEn-eval}

\begin{table}[H]
\centering
\small
\begin{tabular}{M{2.5cm} c c c c}
\toprule
\makecell{Architecture \\+ Matching} & FR	& IT & ES & Avg\\
\toprule
MAST + canonical-aliasing & 100 & 100 & 100 & 100 \\
\toprule
MAST + auto-aliasing	 & 103.0 & 107.1 & 108.4 & 106.0\\
\toprule
MAMT + auto-aliasing & \textbf{104.0} & \textbf{118.4} & \textbf{121.6} & \textbf{113.9}\\
\toprule
\end{tabular}
\caption{Comparison of auto-aliasing and multi-task architecture on Romance languages (numbers are relative to using canonical-aliasing).}
\label{resultsEU:tab}
\end{table}

In this section, we study the applicability of LaTeX-Numeric for three Romance languages. We train a separate (Category-A) model for each Romance language 
replacing English word embeddings with language specific fastText~\cite{conneau2017word} embeddings. Table~\ref{resultsEU:tab} shows results on Test-Romance
dataset. We observe 6.0\% F1 improvement with our proposed auto-aliasing and additional 7.9\% improvement with use of MAMT-NER architecture, showing effectiveness of
our proposed approaches across languages.

\section{Conclusion}

\label{conclusion}
In this paper, we described `LaTeX-Numeric', a high-precision fully-automated framework for training attribute extraction 
models for E-commerce numeric attributes. We characterized the problem of Missing-PA that arises with distant supervision due to missing
attribute values and proposed a multi-task learning architecture to alleviate the Missing-PA problem, leading to 9.2\% F1
improvement for numeric attributes. We established the applicability of our proposed multi-task architecture for textual attributes and BERT as underlying model as well. Additionally, we proposed an automated algorithm for alias creation, to deal with variations of 
numeric attribute mentions, leading to models with 20.2\% F1 improvement. Our evaluation on three Romance languages establishes that these improvements are
applicable across non-English languages as well. Models trained using our proposed LaTeX-Numeric framework achieve high F1 score, making them suitable for practical applications.

\nocite{sogaard2016deep}
\nocite{reimers2017optimal}
\nocite{liu2019multi}

\bibliographystyle{acl_natbib}
\bibliography{anthology,naacl2021}

\end{document}